%% file: scrambling.tex
\documentclass[11pt,a4paper]{article}
\usepackage{etex}
\usepackage{acl2017}
\usepackage{hyperref}
\usepackage{tikz,pgf}
\usepackage{times}
\usepackage{amsmath}
\usepackage{latexsym}
\usepackage{fixltx2e}
\usepackage{scalefnt}
\usepackage{pgfplots}
\usetikzlibrary{fit,backgrounds,shapes,arrows,patterns,pgfplots.groupplots,positioning}
\pgfplotsset{compat=1.5.1}
\usepackage[singlelinecheck=false]{caption}
\usepackage{mdframed}
\usepackage[outercaption]{sidecap}    
\sidecaptionvpos{figure}{t}
\usepackage{breakurl}

\aclfinalcopy 
\def\aclpaperid{4} 

\makeatletter
\renewcommand*{\p@section}{\S} 
\renewcommand*{\p@subsection}{\S}
\renewcommand*{\p@subsubsection}{\S}
\makeatother

\title{Leveraging Newswire Treebanks for Parsing Conversational Data with Argument Scrambling}

\author{Riyaz Ahmad Bhat \\ Department of Computer Science \\
University of Colorado Boulder \\ riyaz.bhat@colorado.edu 
 \And
Irshad Ahmad Bhat \\ LTRC, IIIT-H, Hyderabad \\ Telangana, India \\ irshad.bhat@research.iiit.ac.in
\And
Dipti Misra Sharma \\ LTRC, IIIT-H, Hyderabad \\ Telangana, India \\ dipti@iiit.ac.in}

\date{}

\begin{document}

\maketitle

\begin{abstract}
We investigate the problem of parsing conversational data of morphologically-rich languages such as Hindi where argument scrambling occurs frequently. We evaluate a state-of-the-art non-linear transition-based parsing system on a new dataset containing 506 dependency trees for sentences from Bollywood (Hindi) movie scripts and Twitter posts of Hindi monolingual speakers. We show that a dependency parser trained on a newswire treebank is strongly biased towards the canonical structures and degrades when applied to conversational data. Inspired by Transformational Generative Grammar \cite{chomsky2014aspects}, we mitigate the sampling bias by generating all theoretically possible alternative word orders of a clause from the existing (kernel) structures in the treebank. Training our parser on canonical and transformed structures improves performance on conversational data by around 9\% LAS over the baseline newswire parser.
\end{abstract}

\section{Introduction}
\label{sec:intro}
In linguistics, every language in assumed to have a basic constituent order in a clause \cite{comrie1981language}. In some languages, constituent order is fixed to define the grammatical structure of a clause and the grammatical relations therein, while in others, that convey grammatical information through inflection or word morphology, constituent order assumes discourse function and defines the information structure of a sentence \cite{kiss1995discourse}. Despite word order freedom, most of the morphologically-rich languages exhibit a preferred word order in formal registers such as newswire. Word order alternations are more commonplace in informal language use such as day-to-day conversations and social media content. For statistical parsing, word order alternations (argument scrambling) are a major bottleneck. Given appropriate pragmatic conditions a transitive sentence in a morphologically-rich language allows $n$ factorial ($n$!) permutations, where $n$ is the number of verb arguments and/or adjuncts. Argument scrambling often leads to structural discontinuities. Moreover, these scramblings worsen the data sparsity problem since data-driven parsers are usually trained on a limited size treebank where most of the valid structures may never show up. More importantly, most of the available treebanks are built on newswire text which is more formal \cite{plank2016non}. The chances of any deviation from the canonical word order are smaller, thereby creating sampling bias.\\
\indent A common solution to address the sampling bias is to alter the distribution of classes in the training data by using sampling techniques \citep{van2007experimental}. However, simple sampling techniques such as minority oversampling may not be a feasible solution for parsing argument scramblings which are almost non-existent in the newswire treebanks (see Table \ref{tbl:scram}). Newswire data usually represent only a sample of possible syntactic structures and, therefore, suffer from non-representation of certain classes that encode valid arc directionalities. In the Hindi dependency treebank (HTB) \cite{bhathindi}, for example, dependency relations such as source, time and place are never extraposed. Therefore, we instead generate training examples for varied arc directionalities by transforming the gold syntactic trees in the training data. We experiment with the Hindi dependency treebank and show that such transformations are indeed helpful when we deal with data with diverse word-orders such as movie dialogues. Our work is in conformity with earlier attempts where modifying source syntactic trees to match target distributions benefited parsing of noisy, conversational data \cite{van2009domain,foster2010cba}.

{\small
\begin{SCtable}[][!htb]
\resizebox{0.5\linewidth}{!}{
\begin{tabular}{|c|cc|} \hline
S.No. & Order & Percentage \\ \hline
1 & S O V & 91.83 \\ 
2 & O S V & 7.80 \\ 
3 & O V S & 0.19 \\ 
4 & S V O & 0.19 \\ 
5 & V O S & 0.0 \\ 
6 & V S O & 0.0 \\ \hline
\end{tabular}}
\captionsetup{skip=0.3em, font=scriptsize}
\captionof{table}{\label{tbl:scram} The table shows theoretically possible orders of Subject (S), Object (O) and Verb (V) in transitive sentences in the HTB training data with their percentages of occurrence.}
\end{SCtable}}

\vspace{-1.5em}


\section{Sampling Argument Scrambling via Syntactic Transformations}
In \cite{chomsky2014aspects}, Noam Chomsky famously described syntactic transformations which abstract away from divergent surface realizations of related sentences by manipulating the underlying grammatical structure of simple sentences called kernels. For example, a typical transformation is the operation of subject-auxiliary inversion which generates yes-no questions from the corresponding declarative sentences by swapping the subject and auxiliary positions. These transformations are essentially a tool to explain word-order variations \cite{mahajan1990bar,taylan1984function,king1995configuring}.

We apply this idea of transformations to canonical structures in newswire treebanks for generating trees that represent all of the theoretically viable word-orders in a morphologically-rich language. For example, we create a dependency tree where an indirect object is extraposed by inverting its position with the head verb, as shown in Figure \ref{fig:scram}.

\vspace{1em}
\begin{minipage}{.95\linewidth}
\begin{mdframed}
\begin{minipage}{.42\linewidth}
\input{figures/cannonical.tex}
\end{minipage}
\begin{minipage}{.1\linewidth}
$\Longrightarrow$
\end{minipage}%
\begin{minipage}{.4\textwidth}
\input{figures/scrambling.tex}
\end{minipage}%
\end{mdframed}
\captionsetup{skip=0pt,font=small}
\captionof{figure}{\label{fig:scram} The figure depicts one possible permutation for the sentence \emph{Ram ne Gopal ko kit\={a}b d\={i}.} `Ram ERG Gopal DAT book give.' (Ram gave Gopal a book.). The indirect object \emph{Gopal ko} (red, dashed arcs) is postposed by swapping its position with the ditransitive verb \emph{d\={i}} `give'.}
\end{minipage}
\vspace{.5em}

Recently, a related approach was proposed by \newcite{TACL917}, who employed the concept of creolization to synthesize artificial treebanks from Universal Dependency (UD) treebanks. They transform nominal and verbal projections in each tree of a UD language as per the word-order parameters of other UD language(s) by using their supervised word-order models. In single-source transfer parsing, the authors showed that a parser trained on a target language chosen from a large pool of synthetic treebanks can significantly outperform the same parser when it is limited to selecting from a smaller pool of natural language treebanks. 

\begin{minipage}{.8\linewidth}
\centering
\input{figures/learningcurvesHinParsingFull.tex}
\captionsetup{font=small, skip=.3em}
\captionof{figure}{\label{fig:curves} Learning curves plotted against data size on the X axis and LAS score on the Y axis.}
\end{minipage}
\vspace{.5em}

Unlike \newcite{TACL917}, we do not choose one word-order for a verbal projection based on a target distribution, but instead generate all of its theoretically possible orders. For each dependency tree, we alter the linear precedence relations between arguments of a verbal projection in `$n!$' ways, while keeping their dominance relations intact. However, simply permuting all the nodes of verbal projections can lead to an overwhelming number of trees. For example, a data set of `$t$' syntactic trees each containing an average of 10 nodes would allow around $t\times 10!$ i.e., 3 million possible permutations for our training data size, making training infeasible. Moreover, we may only need a subset of the permutations to have a uniform distribution over the possible word orders. We therefore apply a number of filters to restrict the permutations. First, we only permute a subset of the training data which is representative of the newswire domain. It is often the case that domain specific constructions are covered by a limited number of sentences. This can be seen from the learning curves in Figure \ref{fig:curves}; the learning curves flatten out after 4,000 training instances. Second, for each sentence, we only take the $k$ permutations with the lowest perplexity assigned by a language model where $k$ is set to the number of nodes permuted for each verbal projection. The language model is trained on a large and diverse data set (newswire, entertainment, social media, stories, etc.) Finally, we make sure that the distribution of the possible word-orders is roughly uniform or at least less skewed in the augmented training data.

\section{Evaluation Data}
For an intrinsic evaluation of the parsing models on conversational data, we manually annotated dependency trees for sentences that represent natural conversation with at least one structural variation from the canonical structures of Hindi.\footnote{HTB's conversation section has around $\sim$16,00 sentences taken from fiction which, however, strictly obey Hindi's preferred SOV word-order. Therefore, we needed a new dataset with word-order variations.} We used Bollywood movie scripts as our primary source of conversational data. Although, dialogues in a movie are technically artificial, they mimic an actual conversation. We also mined Twitter posts of Hindi monolingual speakers. Tweets can often be categorized as conversational. The data set was sampled from old and new Bollywood movies and a large set of tweets of Indian language users that we crawled from Twitter using Tweepy\footnote{http://www.tweepy.org/}. For Twitter data, we used an off-the-shelf language identification system\footnote{https://github.com/irshadbhat/litcm} to select Hindi only tweets. From this data, we only want those dialogues/tweets that contain a minimum of one argument scrambling. For this purpose we trained an off-the-shelf convolutional neural network classifier for identifying sentences with argument scrambling \cite{kim:2014:EMNLP2014}.\footnote{https://github.com/yoonkim/CNN\_sentence} We trained the model using the canonical and transformed treebank data and achieved around $\sim$97\% accuracy on canonical and transformed versions of HTB test data.\footnote{The system often misclassified noisy sentences from movie scripts and tweets as scrambled.} After automatic identification, we manually selected 506 sentences from the true positives for annotation. For POS tagging and dependency annotation, we used the AnnCorra guidelines defined for treebanking of Indian languages \cite{bharati2009anncorra}. The data was annotated by an expert linguist with expertise in Indian language treebanking.  The annotations were automatically converted to Universal Dependencies (UD) following UD v1 guidelines for multilingual experimentation \cite{de2014universal}. Table \ref{tbl:soveval} shows the distribution of theoretically possible word orders in transitive sentences in the evaluation set. Unlike their distribution in the HTB training data, the word orders in the evaluation set are relatively less skewed.

\vspace{0.5em}
\resizebox{0.8\linewidth}{!}{
\begin{minipage}{.7\linewidth}
\begin{tabular}{|c|cc|} \hline
S.No. & Order & Percentage \\ \hline
1 & S O V & 33.07 \\ 
2 & O S V & 23.62 \\ 
3 & O V S & 17.32 \\ 
4 & S V O & 14.17 \\ 
5 & V O S & 9.45  \\ 
6 & V S O & 2.36  \\ \hline
\end{tabular}
\end{minipage}
\begin{minipage}{.5\linewidth}
\captionof{table}{\label{tbl:soveval} The table shows theoretically possible orders of Subject, Object and Verb in transitive sentences in the Evaluation set with their percentages of occurrence.}
\end{minipage}}
\vspace{0.5em}

Most of the movie scripts available online and the tweets are written in Roman script instead of the standard Devanagari script, requiring back-transliteration of the sentences in the evaluation set before running experiments. We also need normalization of non-standard word forms prevalent in tweets. We followed the procedure adapted by \newcite{bhat-EtAl:2017:EACLshort} to learn a single back-transliteration and normalization system. We also performed sentence-level decoding to resolve homograph ambiguity in Romanized Hindi vocabulary. 

\section{Experimental Setup}
\label{sec:expsetup}
The parsing experiments reported in this paper are conducted using a non-linear neural network-based transition system which is similar to \cite{kiperwasser2016simple}. The monolingual models are trained on training files of HTB which uses the P\={a}ninian Grammar framework (PG) \cite{bharati1995natural}, while the multilingual models are trained on Universal Dependency Treebanks of Hindi and English released under version 1.4 of Universal Dependencies \cite{11234-1-1827}.

\paragraph{Parsing Models}\enspace Our underlying parsing method is based on the arc-eager transition system \cite{nivre2003efficient}. The arc-eager system defines a set of configurations for a sentence {\tt \small w$_1$,...,w$_n$}, where each configuration {\tt \small C = (S, B, A)} consists of a stack {\tt \small S}, a buffer {\tt \small B}, and a set of dependency arcs {\tt \small A}. For each sentence, the parser starts with an initial configuration where {\tt \small S = [ROOT], B = [w$_1$,...,w$_n$]} and {\tt \small A = $\emptyset$} and terminates with a configuration {\tt \small C} if the buffer is empty and the stack contains the {\tt \small ROOT}. The parse trees derived from transition sequences are given by {\tt \small A}. To derive the parse tree, the arc-eager system defines four types of transitions ({\tt \small $t$}): {\tt \small Shift}, {\tt \small Left-Arc}, {\tt \small Right-Arc}, and {\tt \small Reduce}. 

\begin{table*}[!htb]
\begin{center}
\input{tables/normal_vs_scramble.tex}
\end{center}
\captionsetup{skip=0.1em,font=small,width=14.3cm}
\captionof{table}{\label{tbl:resultspd} Accuracy of our different parsing models on conversational data as well as newswire evaluation sets. Improvements in superscript are over the newswire baseline.}
\end{table*}
\footnotetext{We also experimented with minority oversampling and instance weighting, however improvments over newswire were minimal (see \ref{sec:intro} for possible reasons).}

We use a non-linear neural network to predict the transitions for the parser configurations. The neural network model is the standard feed-forward neural network with a single layer of hidden units. We use 128 hidden units and the {\tt \small RelU} activation function. The output layer uses a softmax function for probabilistic multi-class classification. The model is trained by minimizing negative log-likelihood loss with $l2$-regularization over the entire training data. We use Momentum SGD for optimization \cite{duchi2011adaptive} and apply dropout \cite{hinton2012improving}. 

From each parser configuration, we extract features related to the top three nodes in the stack, the top node in the buffer and the leftmost and rightmost children of the top three nodes in the stack and the leftmost child of the top node in the buffer. Similarly to \newcite{kiperwasser2016simple}, we use two stacked Bidirectional LSTMs with 128 hidden nodes for learning the feature representations over conjoined word-tag sequences for each training sentence. We use an additional Bidirectional LSTM (64 nodes) for learning separate representations of words over their character sequences for capturing out-of-vocabulary (OOV) words at testing time. We use word dropout with a dropout probability of 0.1 which enables character embeddings to drive the learning process around 10\% of the time instead of full word representations. This is important for evaluation on noisy data where OOV words are quite frequent. The monolingual models are initialized using pre-trained 64 dimensional word embeddings of Hindi, while multilingual models use Hindi-English bilingual embeddings from \newcite{bhat-EtAl:2017:EACLshort}\footnote{https://bitbucket.org/irshadbhat/indic-word2vec-embeddings}, while POS embeddings are randomly initialized within a range of {\tt \small -0.25} to {\tt \small +0.25} with 32 dimensions.

Moreover, we use pseudo-projective transformations of \newcite{nivre2005} to handle a higher percentage of non-projective arcs in the evaluation data (6\% as opposed to 2\% in the training data). We use the most informative scheme of {\tt head+path} to store the transformation information. Inverse transformations based on breadth-first search are applied to recover the non-projective arcs in a post-processing step. 

\section{Experiments and Results}
We ran two experiments to evaluate the effectiveness of the tree transformations on the parsing of conversational data. In the first, we leverage the monolingual annotations by applying syntactic transformations; in the second we use a cross-lingual treebank with diverse word-orders. For each experiment type, we report results using both predicted and gold POS tags. The POS taggers are trained using an architecture similar to the parser's with a single layer MLP which takes its input from Bi-LSTM representation of the focus word (see Appendix for the results). We used the newswire parsing models as the baseline for evaluating the impact of tree transformations and multilingual annotations. The augmented models are trained on the union of the original newswire training data and the transformed trees. We generated 9K trees from 4K representative sentences (Figure \ref{fig:curves}) which were projectivized before applying syntactic transformations to preserve non-projective arcs. Our results are reported in Table \ref{tbl:resultspd}. 

As the table shows, our newswire models suffer heavily when applied to conversational data. The parser indeed seems biased towards canonical structures of Hindi. It could not correctly parse extraposed arguments, and could not even identify direct objects if they were not adjacent to the verb. However, in both gold and predicted settings, our augmented parsing models produce results that are approximately 9\% LAS points better than the state-of-the-art baseline newswire parsers \cite{bhat2017improving}. Our augmented models even provided better results with UD dependencies. Probably due to the increased structural ambiguity, augmenting transformed trees with the original training data led to a slight decrease in the results on the original Hindi test sets in both UD and PG dependencies. Interestingly, our cross-lingual model also captured certain levels of scrambling which could be because the English treebank would at least provide training instances for SVO word order. 


\section{Conclusion}
In this paper, we showed that leveraging formal newswire treebanks can effectively handle argument scrambling in informal registers of morphologically-rich languages such as Hindi. Inspired by Chomskyan syntactic tradition, we demonstrated that sampling bias can be mitigated by using syntactic transformations to generate non-canonical structures as additional training instances from canonical structures in newswire. We also showed that multilingual resources can be helpful in mitigating sampling bias.

The~code~of~the~parsing~models~is~available~at~the~GitHub~repository~\burl{https://github.com/riyazbhat/conversation-parser}, while the data can be found under the Universal Dependencies of Hindi at \burl{https://github.com/ UniversalDependencies/UD\_Hindi}.

\bibliographystyle{acl_natbib.bst}
\bibliography{iwpt}

\appendix

\section{Supplementary Material}
\begin{minipage}{.9\linewidth}
\resizebox{.9\columnwidth}{!}{
\input{figures/parser_flow_ft.tex}}
\captionsetup{justification=centering, margin=.001cm}

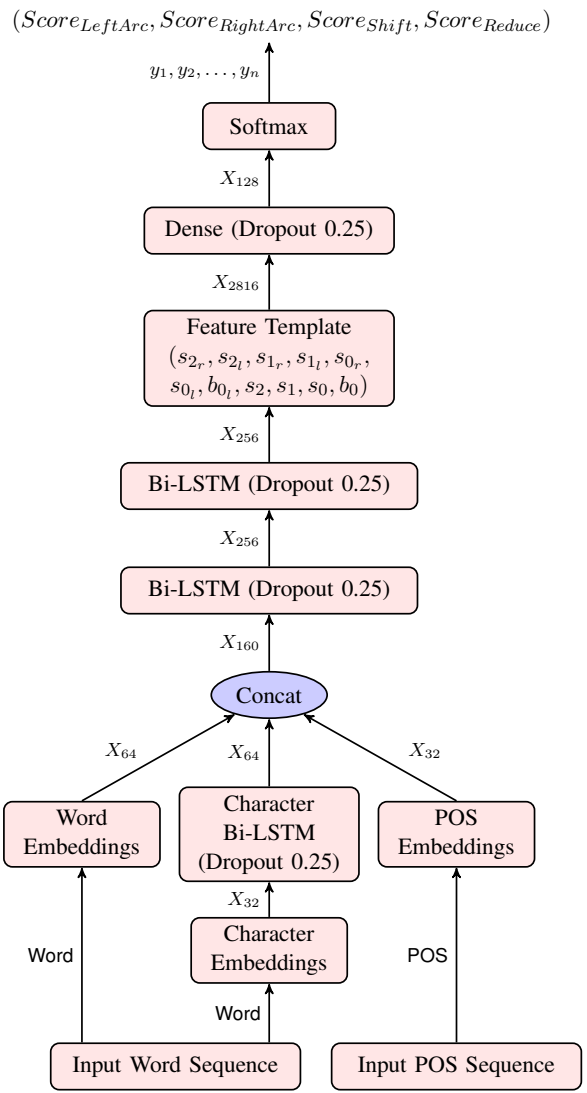
\captionof{figure}{Parsing Architecture}
\end{minipage}

\vspace{.3em}
\begin{minipage}{\linewidth}
\begin{center}
\begin{tabular}{|c|c|c|}\hline
S.No. &Data-set             & Recall \\\hline
1. &Newswire$^{PG}$      & 96.98  \\
2. &Conversation$^{PG}$  & 91.33  \\\hline\hline
3. &Newswire$^{UD}$      & 97.59  \\
4. &Conversation$^{UD}$  & 89.40  \\\hline
\end{tabular}
\captionsetup{width=6cm}
\captionof{table}{POS tagging accuracies on PG and UD evaluation (newswire and converstaion) data.}
\end{center}
\end{minipage}

\end{document}

%% file: figures/cannonical.tex
\resizebox{4.3cm}{!}{
\begin{tikzpicture}[level distance=5em,scale=0.7]
        \node (is-root) {\bf d\={i}}
                [sibling distance=5cm]

                child { node (1){\bf Ram}
                [level distance=1.5cm]
                                child { node[right=.5cm] (a) {\bf ne} }
		}
                child { node (2)[right=.5cm of 1] {\bf Gopal}
                [level distance=1.5cm]
                                child { node[right=.5cm] {\bf ko} }
		}
                child { node (3)[right=1cm of 2] {\bf kit\={a}b} }
                child { node [right=2cm of 3] {.} };
\end{tikzpicture}}

%% file: figures/scrambling.tex
\resizebox{3.3cm}{!}{\begin{tikzpicture}[level distance=5em,scale=0.7]
        \node (is-root) {\bf d\={i}}
                [sibling distance=3cm]

                child { node (1){\bf Ram}
                [level distance=1cm]
                                child { node[right=.7cm] (a) {\bf ne} }
		}
                child { node (2)[right=1cm of 1] {\bf kit\={a}b} }
                child[dashed,red,very thick] {node (3)[right=2cm of 2,black] {\bf Gopal}
                [level distance=1cm]
                                child{ node[right=1cm,black] {\bf ko} }
		}
                child { node [right=1.5cm of 3] {.} };
\end{tikzpicture}}

%% file: figures/learningcurvesHinParsingFull.tex
\pgfplotsset{every tick label/.append style={font=\fontsize{4}{5}\selectfont}}
\begin{tikzpicture}
\begin{axis}[
    width=6.5cm,
    height=4cm,
    title={},
    scaled ticks=false, tick label style={/pgf/number format/fixed},
    xmin=0, xmax=15000,
    ymin=70, ymax=95,
    xtick={0,1000,2000,3000,4000,5000,6000,7000,8000,9000,10000,11000,12000,13000,14000, 15000},
    xticklabels={0,1000,2000,3000,4000,5000,6000,7000,8000,9000,10000,11000,12000,13000,14000,15000},
    x tick label style={rotate=45,anchor=north east, inner sep=0mm},
    ytick={70,75,80,85,90,95,100},
    ymajorgrids=true,
    xmajorgrids=true,
    grid style=dashed,
]
 
\addplot[
    color=blue,
    mark=square,
    mark size=1pt,
    ]
    coordinates {
    (100,74.53)(200,80.34)(300,82.55)(400,84.09)(500,85.64)(600,86.24)(700,86.97)(800,87.03)(900,87.30)(1000,87.76)(2000,89.87)(3000,90.63)(4000,91.57)(5000,91.68)(6000,91.89)(7000,92.00)(8000,92.05)(9000,92.11)(10000,92.17)(11000,92.23)(12000,92.22)(13000,92.31)(14000,92.33)
    };
 
\end{axis}
\end{tikzpicture}

%% file: tables/normal_vs_scramble.tex
\resizebox{.9\linewidth}{!}{
\begin{tabular}{|c|cc|cc||cc|cc||cc|cc|}\cline{1-13}
&\multicolumn{4}{|c||}{\textbf{Newswire$^{PG/UD}$}} & \multicolumn{4}{|c||}{\textbf{Newswire$^{PG/UD}$+Transformed Newswire$^{PG/UD}$}} & \multicolumn{4}{|c|}{\textbf{Newswire$^{UD}$+English$^{UD}$}} \\\cline{2-13}

Data-set&\multicolumn{2}{|c|}{\textbf{Gold POS}} & \multicolumn{2}{|c||}{\textbf{Auto POS}} &\multicolumn{2}{|c|}{\textbf{Gold POS}} & \multicolumn{2}{|c||}{\textbf{Auto POS}} & \multicolumn{2}{|c|}{\textbf{Gold POS}} & \multicolumn{2}{|c|}{\textbf{Auto POS}}\\\cline{2-13}

&                                      UAS   & LAS   &  UAS  &  LAS  & UAS        & LAS        & UAS        &  LAS       &  UAS  &  LAS  &  UAS  & LAS  \\ \hline
\multicolumn{1}{|c|}{Newswire$^{PG}$}       & 96.41 & 92.08 & 94.55 & 89.51 & 96.07$^{-0.34}$ & 91.75$^{-0.33}$ & 94.29$^{-0.26}$ & 89.28$^{-0.23}$ &   -   &   -   &   -   &   -   \\\hline
\multicolumn{1}{|c|}{Conversation$^{PG}$}   & 74.03 & 64.30 & 69.52 & 58.91 & 84.68$^{+10.65}$ & 73.94$^{+9.64}$ & 79.07$^{+9.55}$ & 67.41$^{+8.5}$ &   -   &   -   &   -   &   -   \\\hline 
\multicolumn{1}{|c|}{Newswire$^{UD}$}    & 95.04 & 92.65 & 93.85 & 90.59 & 94.59$^{-0.45}$ & 92.03$^{-0.62}$ & 93.32$^{-0.53}$ & 89.98$^{-0.61}$ & 94.56$^{-0.48}$ & 91.87$^{-0.78}$ & 93.22$^{-0.63}$ & 89.72$^{-0.87}$ \\\hline 
\multicolumn{1}{|c|}{Conversation$^{UD}$}& 73.23 & 64.77 & 68.81 & 59.43 & 83.97$^{+10.74}$ & 74.61$^{+9.84}$ & 78.38$^{+9.57}$ & 67.98$^{+8.55}$ & 77.73$^{+4.5}$ & 68.12$^{+3.35}$ & 71.29$^{+2.48}$ & 62.46$^{+3.03}$ \\\hline 
\end{tabular}}

%% file: figures/parser_flow_ft.tex
%
%

\tikzstyle{ref} = [draw=none, text width=2em, text centered, minimum height=2em]
\tikzstyle{score} = [draw=none, text width=22em, text centered]
\tikzstyle{fblock} = [rectangle, draw, fill=red!10, text width=5em, text centered, rounded corners, minimum height=2em]
\tikzstyle{iblock} = [rectangle, draw, fill=red!10, text width=10em, text centered, rounded corners, minimum height=2em]
\tikzstyle{eblock} = [rectangle, draw, fill=red!10, text width=6em, text centered, rounded corners, minimum height=2em]
\tikzstyle{line} = [draw, -latex']
\tikzstyle{cloud} = [draw, ellipse,fill=blue!20, minimum height=2em]

\begin{minipage}{\linewidth}
\begin{tikzpicture}[->,>=stealth',auto,node distance=3cm,
  thick,main node/.style={circle,draw,font=\sffamily\Large\bfseries}]
    \node [score] (mlpu) {$(Score_{LeftArc}, Score_{RightArc}, Score_{Shift}, Score_{Reduce})$};
    
    \node [fblock, below of=mlpu, node distance=4.5em] (mlpo) {Softmax};

    \node [fblock, below of=mlpo, node distance=4.5em, text width=10em] (mlpd) {Dense (Dropout 0.25)};

    \node [iblock, below of=mlpd, node distance=5.5em] (ft) {Feature Template\\$(s_{2_r}, s_{2_l}, s_{1_r}, s_{1_l}, s_{0_r},$\\
                                                                              $s_{0_l}, b_{0_l}, s_2, s_1, s_0, b_0)$};

    \node [fblock, below of=ft, node distance=5.5em, text width=12em] (bilstm2) {Bi-LSTM (Dropout 0.25)};

    \node [fblock, below of=bilstm2, node distance=4.5em, text width=12em] (bilstm1) {Bi-LSTM (Dropout 0.25)};

    \node [cloud, below of=bilstm1, node distance=4.5em] (concat) {Concat};

    \node [eblock, below of=concat, node distance=6em, text width=7em] (bembc) {Character\\Bi-LSTM\\(Dropout 0.25)};
    \node [eblock, left of=bembc, node distance=8em] (embw) {Word\\Embeddings};
    \node [eblock, right of=bembc, node distance=8em] (embp) {POS\\Embeddings};

    \node [eblock, below of=bembc, node distance=5em] (embc) {Character\\Embeddings};

    \node [ref, below of=embw, node distance=10em] (ip) {};
    \node [ref, left of=ip, node distance=0em] (lipw) {};
    \node [ref, right of=ip, node distance=8em] (ripw) {};
    \node [iblock, right of=ip, node distance=4em] (ipw) {Input Word Sequence};
    \node [iblock, below of=embp, node distance=10em] (ipp) {Input POS Sequence};

    \path[every node/.style={font=\sffamily\small}]

        (mlpo) edge node [left] {$y_1, y_2, \dots, y_n$} (mlpu)
        (mlpd) edge node [left] {$X_{128}$} (mlpo)
        (ft) edge node [left] {$X_{2816}$} (mlpd)
        (bilstm2) edge node [left] {$X_{256}$} (ft)
        (bilstm1) edge node [left] {$X_{256}$} (bilstm2)
        (concat) edge node [left] {$X_{160}$} (bilstm1)
        (embw.north) edge node [left, xshift=-.5em, yshift=0.3em] {$X_{64}$} (concat)
        (bembc) edge node [left] {$X_{64}$} (concat)
        (embp.north) edge node [left, xshift=3em, yshift=0.3em] {$X_{32}$} (concat)
        (embc) edge node [left] {$X_{32}$} (bembc)

        (lipw) edge node [left] {Word} (embw)
        (ripw) edge node [left] {Word} (embc)
        (ipp) edge node [left] {POS} (embp);

\end{tikzpicture}
\end{minipage}

